\documentclass[letterpaper, 10 pt, conference]{ieeeconf}  % Comment this line out if you need a4paper

\IEEEoverridecommandlockouts                              
\pdfoutput=1

\usepackage{graphics} % for pdf, bitmapped graphics files
\usepackage{amsmath} % assumes amsmath package installed
\usepackage{amssymb}  % assumes amsmath package installed
\usepackage{graphicx}
\usepackage{subfigure}
\usepackage{epstopdf}
\usepackage{threeparttable}
\usepackage{multirow}
\usepackage{booktabs}
\usepackage{tabularx}
\usepackage{algpseudocode}
\usepackage{mwe,tikz}
\usepackage[percent]{overpic}
\usepackage{pict2e}

\usepackage{balance}
\usepackage{hyperref}
\usepackage{cleveref}

\newlength{\bibitemsep}
\newlength{\bibparskip}
\let\oldthebibliography\thebibliography
\renewcommand\thebibliography[1]{%
  \oldthebibliography{#1}%
  \setlength{\parskip}{\bibitemsep}%
  \setlength{\itemsep}{\bibparskip}%
}
\DeclareMathAlphabet{\mbf}{OT1}{ptm}{b}{n}
\newcommand{\mbs}[1]{\boldsymbol{#1}}
\newcommand{\mbsbar}[1]{\overline{\boldsymbol{#1}}}
\newcommand{\mbshat}[1]{\hat{\boldsymbol{#1}}}

\newcommand{\mbfbar}[1]{\overline{\mbf{#1}}}

\newcommand{\One}{\mbf{1}}

\DeclareMathAlphabet{\mbfh}{OML}{cmm}{b}{it}

\newcommand{\cframe}[1]{\ensuremath \underrightarrow{\mathcal{F}}_{#1}}

\newcommand{\abs}[1]{\left\vert#1\right\vert}

\newcommand{\bbm}{\begin{bmatrix}}
\newcommand{\ebm}{\end{bmatrix}}

\newcommand\T{\rule{0pt}{2.6ex}}        % Top strut
\newcommand\B{\rule[-1.2ex]{0pt}{0pt}} % Bottom strut

\title{\LARGE \bf
PROBE: Predictive Robust Estimation for Visual-Inertial Navigation
}

\author{Valentin Peretroukhin, Lee Clement, Matthew Giamou, and Jonathan Kelly% <-this % stops a space
\thanks{All authors are at the Institute for Aerospace Studies,  University of Toronto, Canada. \{\tt\small v.peretroukhin, lee.clement, matthew.giamou\}@mail.utoronto.ca, \tt\small jkelly@utias.utoronto.ca.}}

\begin{document}

\maketitle
\thispagestyle{empty}
\pagestyle{empty}

%%%%%%%%%%%%%%%%%%%%%%%%%%%%%%%%%%%%%%%%%%%%%%%%%%%%%%%%%%%%%%%%%%%%%%%%%%%%%%%%
\begin{abstract}
Navigation in unknown, chaotic environments continues to present a significant challenge for the robotics community.
Lighting changes, self-similar textures, motion blur, and moving objects are all considerable stumbling blocks for state-of-the-art vision-based navigation algorithms.
In this paper we present a novel technique for improving localization accuracy within a visual-inertial navigation system (VINS).
We make use of training data to learn a model for the quality of visual features with respect to localization error in a given environment.
This model maps each visual observation from a predefined prediction space of visual-inertial predictors onto a scalar weight, which is then used to scale the observation covariance matrix.
In this way, our model can adjust the influence of each observation according to its quality.
We discuss our choice of predictors and report substantial reductions in localization error on 4 km of data from the KITTI dataset, as well as on experimental datasets consisting of 700 m of indoor and outdoor driving on a small ground rover equipped with a Skybotix VI-Sensor.
\end{abstract}

%%%%%%%%%%%%%%%%%%%%%%%%%%%%%%%%%%%%%%%%%%%%%%%%%%%%%%%%%%%%%%%%%%%%%%%%%%%%%%%%
%%%%%%%%%%%%%%%%%%%%%%%%%%%%%%%%%%%%%%%%%%%%%%%%%%%%%%%%%%%%%%%%%%%%%%%%%%%%%%%%
\section{INTRODUCTION}
Robot navigation relies on an accurate quantification of sensor noise or uncertainty in order to produce reliable state estimates.
In practice, this uncertainty is often fixed for a given sensor and experiment, whether by automatic calibration or by manual tuning.
Although a fixed measure of uncertainty may be reasonable in certain static environments, dynamic scenes frequently exhibit many effects that corrupt a portion of the available observations.
For visual sensors, these effects include, for example, self-similar textures, variations in lighting, moving objects, and motion blur. 
We assert that there may be useful information available in these observations that would normally be rejected by a fixed-threshold outlier rejection scheme. 
Ideally, we would like to retain some of these observations in our estimator, while still placing more trust in observations that do not suffer from such effects.

In this paper we present PROBE, a Predictive ROBust Estimation technique that improves localization accuracy in the presence of such effects by building a model of the uncertainty in the affected visual observations. 
We learn the model in an offline training procedure and then use it online to predict the uncertainty of incoming observations as a function of their location in a predefined \emph{prediction space}.
Our model can be learned in completely unknown environments with frequent or infrequent ground truth data. 

The primary contributions of this research are a flexible framework for learning the quality of visual features with respect to navigation estimates, and a straightforward way to incorporate this information into a navigation pipeline. On its own, PROBE can produce more accurate estimates than a binary outlier rejection scheme like Random Sample Consensus (RANSAC) \cite{Fischler:1981cv} because it can simultaneously reduce the influence of outliers while intelligently weighting inliers. PROBE reduces the need to develop finely-tuned uncertainty models for complex sensors such as cameras, and better accounts for the effects observed in complex, dynamic scenes than typical fixed-uncertainty models. While we present PROBE in the context of visual feature-based navigation, we stress that it is not limited to visual measurements and could also be applied to other sensor modalities.
% The remainder of this paper is organized as follows: In Section \ref{sec:related_work}, we discuss related work in the literature. Section \ref{sec:approach} outlines our tightly-coupled visual-inertial odometry pipeline and learning model. Section \ref{sec:predictors} discusses and justifies our choice of prediction space. Sections \ref{sec:results} and \ref{sec:discussion} discuss experimental results of PROBE's performance on several segments from the KITTI dataset \cite{Geiger:2013kp}, and experimental datasets collected at the University of Toronto Institute for Aerospace Studies (UTIAS). Finally, Section \ref{sec:conclusions} offers some concluding remarks.

\begin{figure}[t]
    \vspace{0.3cm}
    \centering
    \includegraphics[width=0.5\textwidth]{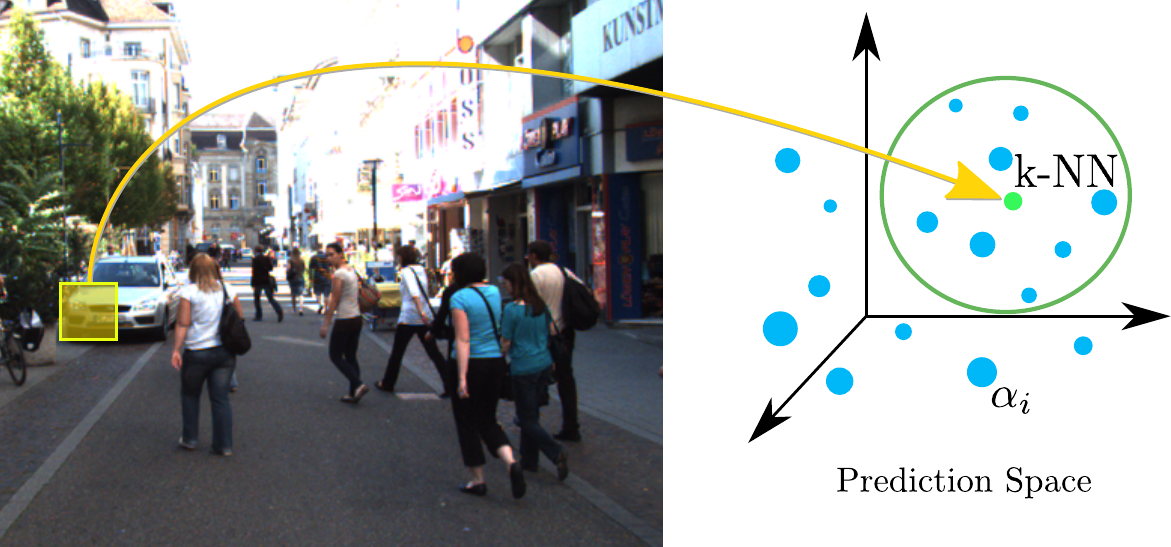}
    \caption{PROBE maps image features into a prediction space to predict feature quality ($\alpha$). Feature quality is a function of the nearest neighbours from training data.}
    \label{fig:feature_space}
    \vspace{-0.25cm}
\end{figure}

%%%%%%%%%%%%%%%%%%%%%%%%%%%%%%%%%%%%%%%%%%%%%%%%%%%%%%%%%%%%%%%%%%%%%%%%%%%%%%%%
%%%%%%%%%%%%%%%%%%%%%%%%%%%%%%%%%%%%%%%%%%%%%%%%%%%%%%%%%%%%%%%%%%%%%%%%%%%%%%%%
\section{RELATED WORK} \label{sec:related_work}
% \section{Related work} \label{sec:related_work}
Two important tasks must be performed to obtain accurate estimates from navigation systems: sensor noise characterization and outlier rejection.
Standard calibration techniques can provide some insight into sensor noise, but in practice, noise parameters are often manually tuned to optimize estimator performance for a given environment.
Outlier rejection typically relies on front-end procedures such as Random Sample Consensus (RANSAC) \cite{Fischler:1981cv, VisualOdometry:n9UHQMsK}, which attempts to remove outliers before the estimator can process them, or back-end procedures such as M-estimation, which uses robust cost functions to limit the influence of large errors \cite{Latif:2014dy}.

For visual systems, a large body of literature has also focused on compensating for specific effects such as moving objects \cite{Wangsiripitak:2009bt}, lighting variations \cite{McManus:2014ew}, and motion blur \cite{Pretto:2009er} using carefully crafted models. %Although such models can work well for specific phenomena, they are limited in their scope and often must rely on parallel computation to achieve online performance.

More general approaches for attempting to select an optimal set of features has also been investigated in the literature. In \cite{Sim:1999ksa}, visual features are parametrized in an attribute space that is used both to establish feature correspondences and to learn a model of feature reliability for pose estimation. Sufficiently reliable features are stored for use as landmarks during pose estimation, while unreliable features are discarded. Likewise, our method maps each feature in a prediction space to a corresponding scalar weight, but we use this weight in our estimator rather than as a criterion to discard data. In more recent work \cite{Carlone:2014dt}, the authors forgo the concept of outliers in favour of finding the largest subset of coherent measurements.  Once again, our method differs in that it reduces the influence of deleterious features without explicitly discarding them.

Our work on predicting the general quality of visual features is inspired by recent research on predictive covariance estimation \cite{VegaBrown:ew}, which estimates full measurement covariances by collecting empirical errors and storing them in a prediction space.
For cameras, learning measurement covariances requires ground truth for the observed quantities, which is difficult to obtain for sparse visual landmarks.
To relax this requirement, the authors of \cite{VegaBrown:2013fv} describe an expectation-maximization technique that can serve as a proxy for ground truth, but it is unclear whether this technique is applicable to sparse vision-based navigation.
In contrast, our method requires ground truth for only a subset of position states and can be straightforwardly applied to standard sparse visual navigation systems.

Finally, adaptive techniques exist that learn scalar weights to intelligently modify the influence of particular measurements within a Kalman filter estimator  \cite{Ting:2007js}. We adopt a similar approach for a visual-inertial navigation pipeline, but do so predictively rather than reactively to respond to new measurements with minimal delay.%  \cite{Sim:1999ksa}, 

%\begin{itemize}
%\item Different ways people deal with dynamic effects
%\item CELLO (Prediction space)
%\item Robust Cost Functions and Dynamic Covariance Scaling
%\item Distraction suppression
%\item Cite myself
%\item \end{itemize}

%%%%%%%%%%%%%%%%%%%%%%%%%%%%%%%%%%%%%%%%%%%%%%%%%%%%%%%%%%%%%%%%%%%%%%%%%%%%%%%%
%%%%%%%%%%%%%%%%%%%%%%%%%%%%%%%%%%%%%%%%%%%%%%%%%%%%%%%%%%%%%%%%%%%%%%%%%%%%%%%%
\section{VISUAL-INERTIAL NAVIGATION SYSTEM} \label{sec:vins}
In this work, we evaluate PROBE as a way to improve navigation estimates from visual-inertial navigation systems (VINS).  VINS have been used onboard many different robotic platforms including micro aerial vehicles (MAVs), ground vehicles, smartphones, and even wearable devices such as Google Glass \cite{Anonymous:m8ztJh1D}. Stereo cameras \cite{Leutenegger:2014hk} and monocular cameras \cite{Anonymous:5tilMnYo, Indelman:2013bca} are common sensor choices for acquiring visual data in such systems.

We choose to implement a stereo VINS adapted from a common stereo visual odometry (VO) framework used in numerous successful mobile robotics applications \cite{Cheng:2006kl, Kelly:2008tr, Geiger:2011jb}.
Unlike other VINS \cite{Leutenegger:2014hk, Anonymous:5tilMnYo}, we choose not to use the linear accelerations reported by the IMU to drive a motion model.
Instead, we opt to use only angular velocities to extract rotation estimates between successive poses. Although this choice reduces the potential accuracy of the method, it also confers several advantages that make our pipeline a solid foundation from which to evaluate PROBE. Specifically, we do not need to keep track of a linear velocity state, accelerometer biases, or orientation relative to gravity. Although seemingly minor, this simplification greatly reduces the complexity of the pipeline, obviates the need for a special starting state that is typically required to make gravity observable, and eliminates several sensitive tuning parameters.

We use the relatively stable rotational rates reported by commercial IMUs to extract accurate relative rotation estimates over short time periods.  Similar to \cite{Kneip:dw}, we bootstrap a crude transformation estimate with the rotation estimate before carrying out a full non-linear point cloud alignment.

%(typical rotational drift rates are an the order of 0.1$^\circ$ per minute \cite{Leutenegger:2014hk}) 
Finally, because we assume a stereo camera, the metric scale of our solution is directly observable from the visual measurements, whereas monocular VINS must rely on noisy linear accelerations to determine metric scale \cite{Kelly:2011bw}.

\subsection{Observation Model}
To begin, we outline our point feature observation model. We define two coordinate frames, $\cframe{a}$ and $\cframe{b}$, which represent the stereo camera pose at times $t_a$ and $t_b$ (where $t_b > t_a$), respectively. The coordinates $\mbf{p}_a$ of a landmark observed in $\cframe{a}$ can be transformed into its coordinates $\mbf{p}_b$ in $\cframe{b}$ as follows:
\begin{equation}
	\mbf{p}_b =  \mbf{C}_{ba}(\mbf{p}_a - \mbf{r}_a^{ba}).
\end{equation}
$\mbf{r}_a^{ba}$ is a vector from the origin of $\cframe{a}$ to the origin of $\cframe{b}$ expressed in $\cframe{a}$, and $\mbf{C}_{ba}$ is the rotation matrix from $\cframe{a}$ to $\cframe{b}$.

Assuming rectified stereo images, point $\mbf p$ is projected from the camera frame (with its origin in the centre of the left camera) into image coordinates $\mbf y$ in the left and right cameras according to the camera model
%CHANGE THIS TO LEFT CAMERA
\begin{gather}
\label{eqn:imgCoord}
\mbf{y} = \mbf f (\mbf p) =  \bbm u_l \\ v_l \\ u_r \\  v_r \ebm = \frac{f}{z}
        \bbm x \\  y \\ x -  b \\  y \\
      \ebm
      +
      \bbm
      c_u \\ c_v \\ c_u \\ c_v
      \ebm.
\end{gather}

\noindent Here, $(x,y,z)$ are the components of $\mbf p$, $(u_l,u_r,v_l,v_r)$ are the horizontal and vertical pixel coordinates in each camera respectively, $b$ is the baseline of the stereo pair, $(c_u,c_v)$ is the principal point, and $f$ is the focal length.
\subsection{Direct Motion Solution}

Assuming all landmarks are tracked correctly, our goal is to calculate the transformation from $\cframe{a}$ to $\cframe{b}$, parametrized by $\mbf{C}_{ba}$ and $\mbf{r}_a^{ba}$.
We obtain an estimate for $\mbf{C}_{ba}$, (denoted by $\overline{\mbf{C}}_{ba}$) by integrating angular velocities, $\mbs \omega_j$, from the IMU that are recorded between $t_a$ and $t_b$. 
Defining $\mbs \psi_j := \mbs \omega_j \Delta t$, $\psi_j := \abs{\mbs \psi_j}$, and $\mbshat \psi_j := \mbs \psi_j / \psi_j$, we have
\begin{align}
	\mbs \Psi_j &= \cos \psi_j \mbf 1 + (1 - \cos \psi_j) \mbshat \psi_j \mbshat \psi_j^T - \sin \psi_j \mbshat \psi_j^\times, \\
   \overline{\mbf{C}}_{ba} &= \mbf{C}_{cv} \mbs \Psi_{J} \cdots \mbs \Psi_2 \mbs \Psi_1 \mbf{C}_{cv}^T
\end{align}
where $J$ is the number of IMU measurements between $t_a$ and $t_b$, $\mbf {C}_{cv}$ is the rotation from the IMU frame to the camera frame, $\One$ is the identity matrix, and $(\cdot)^\times$ is the skew-symmetric cross-product operator. Given our rotation estimate, we can compute an estimate for translation as
\begin{equation} \label{eqn:cab_umeyama}
\overline{\mbf{r}}_a^{ba} = -\overline{\mbf{C}}_{ba}^T \mbf{u}_b + \mbf{u}_a,
\end{equation}
where $\mbf{u}_a, \mbf{u}_b$ are the point cloud centroids given by
\begin{align}
\mbf{u}_a &= \frac{1}{N} \sum_{i=1}^{N} \mbf{p}_a^i & \text{and} && \mbf{u}_b &= \frac{1}{N} \sum_{i=1}^{N} \mbf{p}_b^i .
\end{align}

\subsection{Non-linear Motion Solution}

The direct motion solution forms an initial guess \{$\overline{\mbf{C}}_{ba}$, $\overline{\mbf{r}}_a^{ba}$\} for a non-linear optimization procedure. 
Given $N$ tracked points, we wish to minimize a 3D point cloud alignment error
\begin{align} \label{eqn:matrixOptfun}
\mathcal{L} = \frac{1}{2} \sum_{i=1}^{N} & \left( \mbf{p}_b^i - \mbf{C}_{ba}(\mbf{p}_a^i - \mbf{r}_a^{ba}) \right)^T \notag \\
                                        & \mbs{\Gamma}^i \left( \mbf{p}_b^i - \mbf{C}_{ba}(\mbf{p}_a^i - \mbf{r}_a^{ba}) \right),
\end{align}
with
\begin{equation} \label{eqn:gamma}
\mbs{\Gamma}^i = \left( \mbf{G}_b^i \mbf{R}_b^{i} \mbf{G}_b^{i^T} + \mbf{C}_{ba} \mbf{G}_a^i \mbf{R}_a^{i^T} \mbf{G}_a^{i^T} \mbf{C}_{ba}^T   \right)^{-1},
\end{equation}
where
\begin{align}
\mbf{G}_a^i &= \frac{\partial \mbf{g} }{\partial \mbf y }\bigg|_{\mbf{f}(\mbf{p}^i_a)} & \text{and} && \mbf{G}_b^i &= \frac{\partial \mbf{g} }{\partial \mbf y }\bigg|_{\mbf{f}(\mbf{p}^i_b)}
\end{align}
are the Jacobians of the inverse camera model $\mbf{g} := \mbf{f}^{-1}$, and $\mbf{R}_a^{i}$, $\mbf{R}_b^{i}$ are the covariance matrices of the $i^\text{th}$ point in image space. 

We proceed by perturbing the initial guess by a vector $\mbs \xi = \bbm\mbs{\epsilon}^T & \mbs{\phi}^T \ebm^T$, where $\mbs{\epsilon}$ is a translational perturbation and $\mbs{\phi}$ is a rotational perturbation:
\begin{align}
\mbf{r}_a^{ba} &= \mbf{\overline{r}}_a^{ba} + \mbs{\epsilon}, \label{eqn:trans_perturb} \\
\mbf{C}_{ba} &= e^{-\mbs{\phi}^{\times}} \mbfbar{C}_{ba} \approx (\mbf{1} - \mbs{\phi}^{\times})\mbfbar{C}_{ba}. \label{eqn:rot_perturb}
\end{align}

Inserting \eqref{eqn:trans_perturb} and \eqref{eqn:rot_perturb} into \eqref{eqn:matrixOptfun}, we arrive at a cost function that is quadratic in the perturbations:
\begin{gather}
\label{eqn:matrixOptfunPerturbed}
\mathcal{L} \approx \frac{1}{2} \sum_{j=1}^{N} \left( \mbfbar{e}^i + \mbfbar{E}^i \mbs{\xi} \right)^T \mbsbar{\Gamma}^i \left(  \mbfbar{e}^i + \mbfbar{E}^i \mbs{\xi} \right),
\end{gather}
\noindent where
\begin{align*}
\mbfbar{e}^i &= \mbf{p}_b^i - \mbfbar{C}_{ba}(\mbf{p}_a^i - \mbfbar{r}_a^{ba}), \\  \mbfbar{E}^i &= \left[\mbfbar{C}_{ba} ~ -\left(\mbfbar{C}_{ba}(\mbf{p}_a^i - \mbfbar{r}_a^{ba})\right)^\times \right],
\end{align*}
and $\mbsbar \Gamma^i$ indicates that $\mbfbar{C}_{ba}$ has replaced $\mbf{C}_{ba}$ in \eqref{eqn:gamma}.
Setting the derivative of \eqref{eqn:matrixOptfunPerturbed} with respect to $\mbs \xi$ to zero yields a system of linear equations in the optimal update step $\mbs \xi^*$:
\begin{gather}
\label{eqn:matrixOptLinEqns}
\sum_{j=1}^{N} \left( \mbfbar{E}^{i^T} \mbsbar{\Gamma}^i \mbfbar{E}^i \right) \mbs \xi^* = - \sum_{i=1}^{N} \mbfbar{E}^{i^T} \mbsbar{\Gamma}^i \mbfbar{e}^i.
\end{gather}

\noindent Once $\mbs \xi^*$ is determined, the state estimate can be repeatedly updated using:
\begin{gather}
\mbfbar{C}_{ba} \leftarrow \exp((-\mbs{\phi^*})^{\times}) \mbfbar{C}_{ba}, \\
\mbfbar{r}_a^{ba} \leftarrow \mbfbar{r}_a^{ba} + \mbs{\epsilon^*}.
\end{gather}

\noindent In practice, \eqref{eqn:matrixOptLinEqns} is also modified with Levenberg-Marquardt damping to improve convergence.

%%%%%%%%%%%%%%%%%%%%%%%%%%%%%%%%%%%%%%%%%%%%%%%%%%%%%%%%%%%%%%%%%%%%%%%%%%%%%%%%
%%%%%%%%%%%%%%%%%%%%%%%%%%%%%%%%%%%%%%%%%%%%%%%%%%%%%%%%%%%%%%%%%%%%%%%%%%%%%%%%

\section{PROBE: PREDICTIVE ROBUST ESTIMATION} \label{sec:probe}
The aim of PROBE is to learn a model for the quality of visual features, with the goal of reducing the impact of deleterious visual effects such as moving objects, motion blur, and shadows on navigation estimates.
Feature quality is characterized by a scalar weight, $\beta_i$, for each visual feature in an environment.
To compute $\beta_i$ we define a prediction space (similar to \cite{VegaBrown:ew}) that consists of a set of visual-inertial predictors computed from the local image region around the feature and the inertial state of the vehicle (Section \ref{sec:predictors} details our choice of predictors).
We then scale the image covariance of each feature ($\mbf{R}_a^{i}$, $\mbf{R}_b^{i}$ in \eqref{eqn:gamma}) by $\beta_i$ during the non-linear optimization.

In a similar manner to M-estimation, PROBE achieves robustness by varying the influence of certain measurements.
However, in contrast to robust cost functions that weight measurements based purely on estimation error, PROBE weights measurements based on their assessed quality.

To learn the model, we require training data that consists of a traversal through a typical environment with some measure of ground truth for the path, but not for the visual features themselves. Like many machine learning techniques, we assume that the training data is representative of the test environments in which the learned model will be used. 

We learn the quality of visual features \textit{indirectly} through their effect on navigation estimates. We define high quality features as those that result in estimates that are close to ground truth. Our framework is flexible enough that we do not require ground truth at every image and we can learn the model based on even a single loop closure error.

\subsection{Training}
Training proceeds by traversing the training path, selecting a subset of visual features at each step, and using them to compute an incremental position estimate. By comparing the estimated position to the ground truth position, we compute the translational Root Mean Squared Error (RMSE), denoted by $ \alpha_{l,s} $ for iteration $l$ and step $s$, and store it at each feature's position in the prediction space (we denote the set of predictors and associated RMSE value by $\Theta_{l,s}$). The full algorithm is summarized in Figure \ref{fig:ProbeTraining}. Note that $\alpha_{l,s}$ can be computed at each step, at intermittent steps, or for an entire path, depending on the availability of ground truth data.
\begin{figure}[h]
\begin{algorithmic}[1]
\Procedure{trainPROBEModel}{}
\For{$l\gets 1, totalLearningIterations$}
\For{$s\gets 1, totalPathSteps$}
	\State $f_1, \dots, f_J \gets visualFeatureSubset(l)$
	\State $\mbs\pi^1_l, \dots, \mbs\pi^J_l \gets predictors(f_1, \dots, f_J)$
	\State $\bar{\mbf{C}}_{ba},  \bar{\mbf{r}}_a^{ba} \gets poseChange(f_1, \dots, f_J)$
	\State $\mbf \alpha_{l,s}  \gets computeRMSE(\bar{\mbf{r}}_a^{ba}, {\mbf{r}_{a}^{ba}}_{GT} )$
	\State  $ \Theta_{l,s} \gets \left\{\mbs\pi^1_{l,s}, \dots, \mbs\pi^J_{l,s}, \alpha_{l,s}\right\}$
\EndFor
\EndFor
\State \textbf{return} $\mbs \Theta = \{ \Theta_{l,s} \}$
\EndProcedure
\end{algorithmic}
\caption{The PROBE training procedure.}\label{fig:ProbeTraining}
\end{figure}
	\subsection{Evaluation}
To use the PROBE model in a test environment, we compute the location of each observed visual feature in our prediction space, and then compute its relative weight $\beta_i$ as a function of its $K$ nearest neighbours in the training set.
For efficiency, the $K$ nearest neighbours are found using a $k$-d tree.
The final scaling factor $\beta_i$ is a function of the mean of the $\alpha$ values corresponding to the $K$ nearest neighbours, normalized by $\overline \alpha$, the mean $\alpha$ value of the entire training set.

\begin{figure}[h]
\begin{algorithmic}[1]
\Procedure{usePROBEModel}{$\mbs \Theta$}
\For{$i \gets 1, totalFeatures$}
	\State $\mbs \pi_i  \gets predictors(f_i) $
	\State $\alpha_1,...,\alpha_K \gets findKNN(\mbs \pi_i, K, \mbs\Theta)$
	\State $\beta_i \gets \left(\frac{1}{\overline{\alpha} K} \sum_{k=1}^K \alpha_k  \right)^{\gamma}$
\EndFor
\State \textbf{return}  $\mbs \beta = \{\beta_i\}$
\EndProcedure
\end{algorithmic}
\caption{The PROBE evaluation procedure.}\label{fig:ProbeTest}
\vspace{-1em}
\end{figure}

The value of $K$ can be determined through cross-validation, and in practice depends on the size of the training set and the environment.
The computation of $\beta_i$ is designed to map small differences in learned $\alpha$ values to scalar weights that span several orders of magnitude.
An appropriate value of $\gamma$ can be found by searching through a set range of candidate values and choosing the value that minimizes the average RMSE (ARMSE) on the training set.

\subsection{Prediction Space} \label{sec:predictors}
A crucial component of our technique is the choice of prediction space.
In practice, feature tracking quality is often degraded by a variety of effects such as motion blur, moving objects, and textureless or self-similar image regions.
% Features suffering from such effects are often poorly localized in the image, yet may contain useful information that we would like to retain in our estimator.
% Conversely, we would like to place more trust in features that do not suffer from such effects.
The challenge is in determining predictors that account for such effects without requiring excessive computation.
In our implementation, we use the following predictors, but stress that the choice of predictors can be tailored to suit particular applications and environments:
\begin{itemize}
    \item Angular velocity and linear acceleration magnitudes
    \item Local image entropy
    \item Blur (quantified by the blur metric of \cite{Anonymous:Ngi3VEEU})
    \item Optical flow variance score
    \item Image frequency composition
\end{itemize}
We discuss each of these predictors in turn.

\subsubsection{Angular velocity and linear acceleration}
While most of the predictors in our system are computed directly from image data, the magnitudes of the angular velocities and linear accelerations reported by the IMU are in themselves good predictors of image degradation (e.g., image blur) and hence poor feature tracking. 
%We do not explicitly correct for bias in linear accelerations because we expect real motion-induced acceleration to trump bias at the timescales of our test trials.  As a result, there is virtually no computational cost involved in incorporating these quantities as predictors.

\subsubsection{Local image entropy}
Entropy is a statistical measure of randomness that can be used to characterize the texture in an image or patch.
Since the quality of feature detection is strongly influenced by the strength of the texture in the vicinity of the feature point, we expect the entropy of a patch centered on the feature to be a good predictor of its quality.
We evaluate the entropy $S$ in an image patch by sorting pixel intensities into $N$ bins and computing
\begin{equation}
    S = -\sum_{i=1}^N c_i \log_2(c_i),
\end{equation}
where $c_i$ is the number of pixels counted in the $i^\text{th}$ bin.

\subsubsection{Blur}
\begin{figure}
    \centering
    \includegraphics[width=0.4\textwidth]{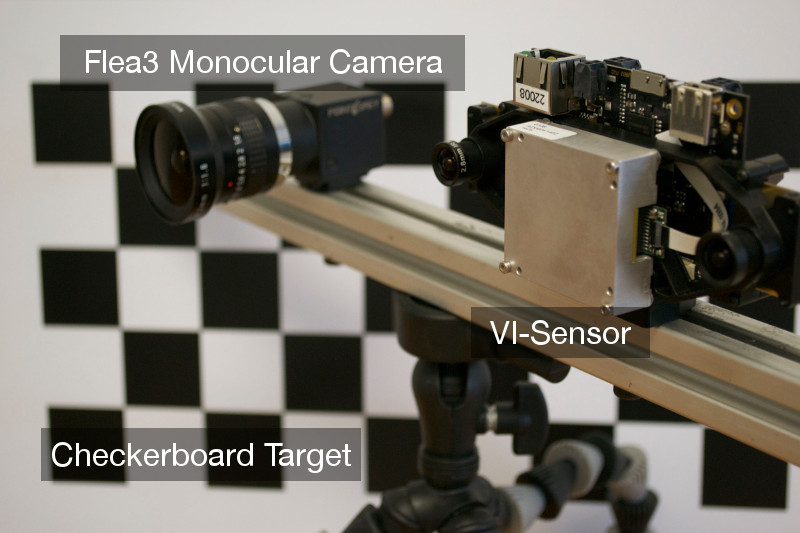}
    \caption{The Skybotix VI-Sensor, Point Grey Flea3, and checkerboard target used in our motion blur experiments.}
    \label{fig:tricifix}
\end{figure}

Blur can arise from a number of sources including motion, dirty lenses, and sensor defects.
All of these have deleterious effects on feature tracking quality.
To assess the effect of blur in detail, we performed a separate experiment.
We recorded images of 32 interior corners of a standard checkerboard calibration target using a low frame-rate (20 FPS) Skybotix VI-Sensor stereo camera and a high frame-rate (125 FPS) Point Grey Flea3 monocular camera rigidly connected by a bar (Figure \ref{fig:tricifix}).
Prior to the experiment, we determined the intrinsic and extrinsic calibration parameters of our rig using the Kalibr package \cite{Furgale:2013dm}.
The apparatus underwent both slow and fast translational and rotational motion, which induced different levels of motion blur as quantified by the blur metric proposed by \cite{Anonymous:Ngi3VEEU}.
% The two motion regimes induced different levels of motion blur, which we distinguish by thresholding the blur metric proposed by \cite{Anonymous:Ngi3VEEU}.

%\begin{figure}
%    \centering
%    \includegraphics[width=0.4\textwidth]{figs/blurMetric_short}
%    \caption{Blur metric \cite{Anonymous:Ngi3VEEU} computed for the left camera of the VI-Sensor for the checkerboard dataset. We separate the dataset into regions of high and low blur corresponding to fast and slow motion, respectively.}
%    \label{fig:visensor_blurMetric}
%\end{figure}

\begin{figure*}
    \centering
    \subfigure[Reprojection error of checkerboard corners triangulated from the VI-Sensor and reprojected into the Flea3.] {
        \includegraphics[width=0.45\textwidth]{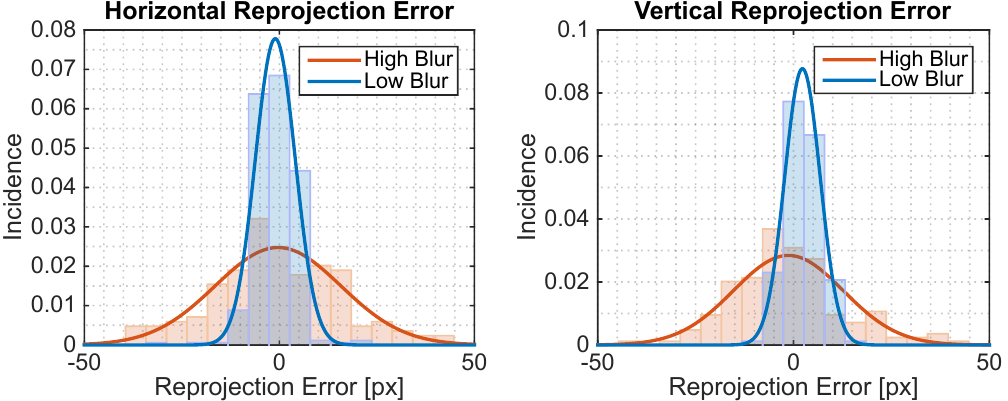}
        \label{fig:visensor_reprojectionError}
    }
    ~
    \subfigure[Tracking error of KLT-tracked checkerboard corners in the left VI-Sensor camera compared to directly re-detected corners.] {
        \includegraphics[width=0.45\textwidth]{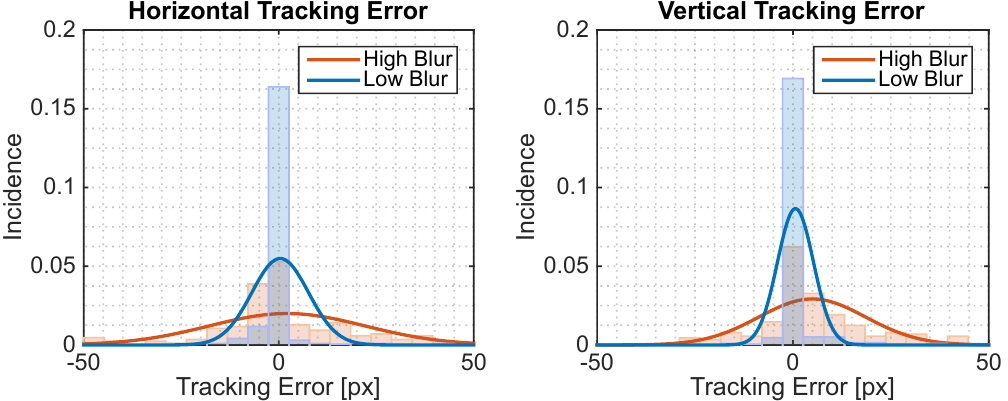}
        \label{fig:visensor_trackingError}
    }
    \caption{Effect of blur on reprojection and tracking error for the slow-then-fast checkerboard dataset. We distinguish between high and low blur by thresholding the blur metric \cite{Anonymous:Ngi3VEEU}. The variance in both errors increases with blur.}
    \label{fig:visensor_histograms}
\end{figure*}

We detected checkerboard corners in each camera at synchronized time steps, computed their 3D coordinates in the VI-Sensor frame, then reprojected these 3D coordinates into the Flea3 frame.
We then computed the reprojection error as the distance between the reprojected image coordinates and the true image coordinates in the Flea3 frame.
Since the Flea3 operated at a much higher frame rate than the VI-Sensor, it was less susceptible to motion blur and so we treated its observations as ground truth.
We also computed a tracking error by comparing the image coordinates of checkerboard corners in the left camera of the VI-Sensor computed from both KLT tracking \cite{Lucas:1981uw} and re-detection.

Figure \ref{fig:visensor_histograms} shows histograms and fitted normal distributions for both reprojection error and tracking error.
From these distributions we can see that the errors remain approximately zero-mean, but that their variance increases with blur.
This result is compelling evidence that the effect of blur on feature tracking quality can be accounted for by scaling the feature covariance matrix by a function of the blur metric.

\subsubsection{Optical flow variance score}
To detect moving objects, we compute a score for each feature based on the ratio of the variance in optical flow vectors in a small region around the feature to the variance in flow vectors of a larger region.
Intuitively, if the flow variance in the small region differs significantly from that in the larger region, we might expect the feature in question to belong to a moving object, and we would therefore like to trust the feature less.
Since we consider only the variance in optical flow vectors, we expect this predictor to be reasonably invariant to scene geometry.

\begin{figure}
    \centering
    \includegraphics[width=0.4\textwidth]{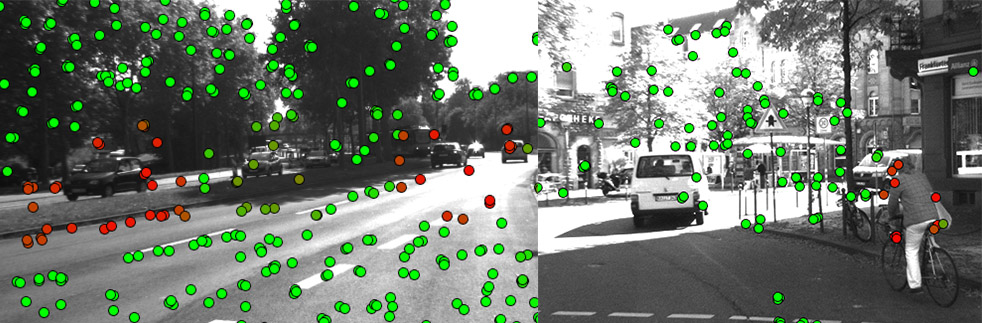}
    \caption{The optical flow variance predictor can help in detecting moving objects. Red circles correspond to higher values of the optical flow variance score (i.e., features more likely to belong to a moving object).}
    \label{fig:flow_variance}
\end{figure}

We compute this optical flow variance score according to
\begin{equation}
    \log \left( \frac{\bar{\sigma}^2_s}{\bar{\sigma}^2_l} \right),
\end{equation}
where $\bar{\sigma}^2_s, \bar{\sigma}^2_l$ are the means of the variance of the vertical and horizontal optical flow vector components in the small and large regions respectively.
Figure \ref{fig:flow_variance} shows sample results of this scoring procedure for two images in the KITTI dataset \cite{Geiger:2013kp}.
Our optical flow variance score generally picks out moving objects such as vehicles and cyclists in diverse scenes.

\subsubsection{Image frequency composition}
\begin{figure}
    \centering
    \includegraphics[width=0.4\textwidth]{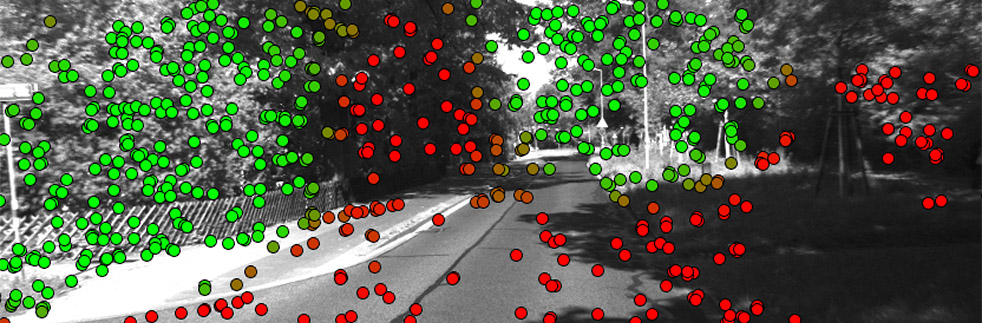}
    \caption{A high-frequency predictor can distinguish between regions of high and low texture such as foliage and shadows. Green indicates higher values.}
    \label{fig:high_frequency}
\end{figure}

Reliable feature tracking is often difficult in textureless or self-similar environments due to low feature counts and false matches.
We detect textureless and self-similar image regions by computing the Fast Fourier Transform (FFT) of each image and analyzing its frequency composition.
For each feature, we compute a coefficient for the low- and high-frequency regimes of the FFT.
Figure \ref{fig:high_frequency} shows the result of the high-frequency version of this predictor on a sample image from the KITTI dataset \cite{Geiger:2013kp}.
Our high-frequency predictor effectively distinguishes between textureless regions (e.g., shadows and roads) and texture-rich regions (e.g., foliage).

%%%%%%%%%%%%%%%%%%%%%%%%%%%%%%%%%%%%%%%%%%%%%%%%%%%%%%%%%%%%%%%%%%%%%%%%%%%%%%%%
%%%%%%%%%%%%%%%%%%%%%%%%%%%%%%%%%%%%%%%%%%%%%%%%%%%%%%%%%%%%%%%%%%%%%%%%%%%%%%%%

\section{EXPERIMENTAL RESULTS} \label{sec:results}

\begin{table*}
    \centering
    \caption{Comparison of translational Average Root Mean Square Error (ARMSE) and Final Translational Error on the KITTI dataset.}

    % Need to add an asterisk to the 30_drive_0027 "Type" to indicate that it's trained on City data, but can't do it without messing up the justification - Matt
    \begin{threeparttable}
        \begin{tabular}{cccccccccccc}
        & & & & \multicolumn{2}{c}{Nominal RANSAC} & & \multicolumn{2}{c}{Aggressive RANSAC} & & &  \\
        	 & & & & \multicolumn{2}{c}{(99\% outlier rejection)} & & \multicolumn{2}{c}{(99.99\% outlier rejection)} & & \multicolumn{2}{c}{PROBE} \B \\
             \cline{5-6} \cline{8-9} \cline{11-12}
        	 Trial & Type & Path Length &&  ARMSE & Final Error & & ARMSE & Final Error & & ARMSE & Final Error \T\B \\ \hline \T
        	\texttt{26\_drive\_0051} & City \tnote{1} & 251.1 m && 4.84 m & 12.6 m && 3.30 m & 8.62 m && 3.48 m & 8.07 m \\
        	\texttt{26\_drive\_0104} & City \tnote{1} & 245.1 m && 0.977 m & 4.43 m && 0.850 m & 3.46 m && 1.19 m & 3.61 m \\ 
        	\texttt{29\_drive\_0071} & City \tnote{1} & 234.0 m && 5.44 m & 30.3 m && 5.44 m & 30.4 m && 3.03 m & 12.8 m \\ 
        		\texttt{26\_drive\_0117} & City \tnote{1} & 322.5 m && 2.29 m & 9.07 m && 2.29 m & 9.07 m && 2.76 m & 9.08 m \\ 
        		\texttt{30\_drive\_0027} & Residential \tnote{1, \dag} & 667.8 m && 4.22 m & 12.2 m && 4.30 m & 10.6 m && 3.64 m & 4.57 m \\ 
        		\texttt{26\_drive\_0022} & Residential \tnote{2} & 515.3 m && 2.21 m & 3.99 m && 2.66 m & 6.09 m && 3.06 m & 4.99 m \\ 
        		\texttt{26\_drive\_0023} & Residential \tnote{2} & 410.8 m && 1.64 m & 8.20 m && 1.77 m & 8.27 m && 1.71 m & 8.13 m \\ 
        		\texttt{26\_drive\_0027} & Road \tnote{3} & 339.9 m && 1.63 m & 8.75 m && 1.63 m & 8.65 m && 1.40 m & 7.57 m \\ 
        		\texttt{26\_drive\_0028} & Road \tnote{3} & 777.5 m && 4.31 m & 16.9 m && 3.72 m & 13.1 m && 3.92 m & 13.2 m \\ 
        		\texttt{30\_drive\_0016} & Road \tnote{3} & 405.0 m && 4.56 m & 19.5 m && 3.33 m & 14.6 m && 2.76 m & 13.9 m \\ 
                UTIAS Outdoor & Snowy parking lot & 302.0 m && 7.24 m & 10.1 m && 7.02 m & 10.6 m && 6.85 m & 6.09 m \\ 
                UTIAS Indoor & Lab interior & 32.83 m && --- & 0.854 m && --- & 0.738 m && --- & 0.617 m  \B \\    
        \hline
            \label{table:kitti_data}
        \end{tabular}
        \begin{tablenotes}
            \item[1] Trained using sequence \texttt{09\_26\_drive\_0005}. ~$^2$ Trained using sequence  \texttt{09\_26\_drive\_0046}. ~$^3$ Trained using sequence  \texttt{09\_26\_drive\_0015}.  
            \item[\dag] This residential trial was evaluated with a model trained on a sequence from the city category because of several moving vehicles that were better represented in that training dataset.
        \end{tablenotes}
    \end{threeparttable}
\end{table*}

\subsection{Datasets}
\begin{figure*}
    \centering
    \includegraphics[width=0.9\textwidth]{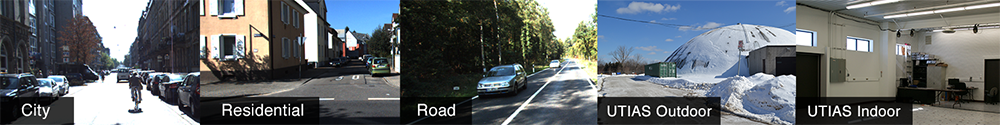}
    \caption{Three types of environments in the KITTI dataset, as well as 2 types of environments at the University of Toronto.  We use one trial from each category to train and then evaluate separate trials in the same category.}
    \label{fig:KITTI-Types}
    \vspace{-0.2cm}
\end{figure*}

\begin{figure}
    \centering
    \includegraphics[width=0.4\textwidth]{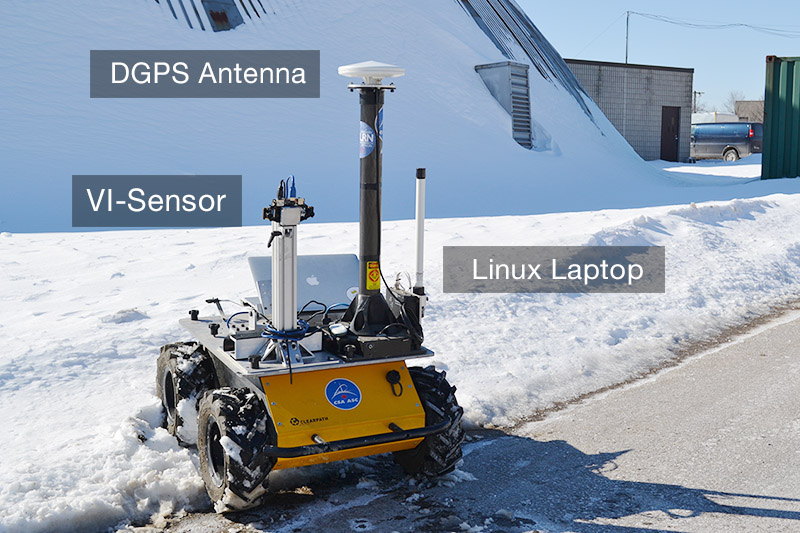}
    \caption{Our four-wheeled skid-steered Clearpath Husky rover equipped with Skybotix VI-Sensor and Ashtech DGPS antenna used to collect the outdoor UTIAS dataset.}
    \label{fig:huskypic}
\end{figure}

% Do you mean train or train and evaluate? The evaluation trials only add up to 4.17 km - Matt
We trained and evaluated PROBE in two sets of experiments.
The first set of experiments made use of 4.5 km of data from the City, Residential, and Road categories of the KITTI dataset \cite{Geiger:2013kp}.
In the second set of experiments, we collected indoor and outdoor datasets at the University of Toronto Institute for Aerospace Studies (UTIAS) using a Skybotix VI-Sensor mounted on an Adept MobileRobots Pioneer 3-AT rover and a Clearpath Husky rover, respectively.
In both cases, the camera recorded stereo images at 10 Hz while the IMU operated at 200 Hz.
The outdoor dataset consisted of a 264 m training run followed by a 302 m evaluation run, with ground truth provided by RTK-corrected GPS.
The indoor dataset consisted of a 32 m training run and a 33 m evaluation run through a room with varying lighting and shadows.
For the indoor dataset, no ground truth was available, so we trained PROBE using only the knowledge that the training path should form a closed loop.

We compare PROBE to what we call the nominal VINS, as well as a VINS with an aggressive RANSAC routine. In the nominal pipeline, we use RANSAC with enough iterations to be 99\% confident that we select only inliers when as many as 50\% of the features are outliers. In the aggressive case, we increase the confidence level to 99.99\%. When PROBE is used, we apply a pre-processing step that makes use of the rotational estimate from the IMU to reject any egregious feature matches by thresholding the cosine distance between pairs of matched feature vectors. We assume small translations between frames and typically set the threshold to reject feature vectors that are separated by more than five degrees. 

For feature extraction, matching, and sparse optical flow calculations, we use the open source vision library \texttt{LIBVISO2} \cite{Geiger:2011jb}.
\texttt{LIBVISO2} efficiently detects thousands of feature correspondences by finding stable features using blob and corner masks and matching them by comparing Sobel filter responses.
For all prediction space calculations, we use features in the left image of the stereo pair.

\subsection{Training}
\begin{figure}
    \centering
    \begin{overpic}[width=0.4\textwidth]{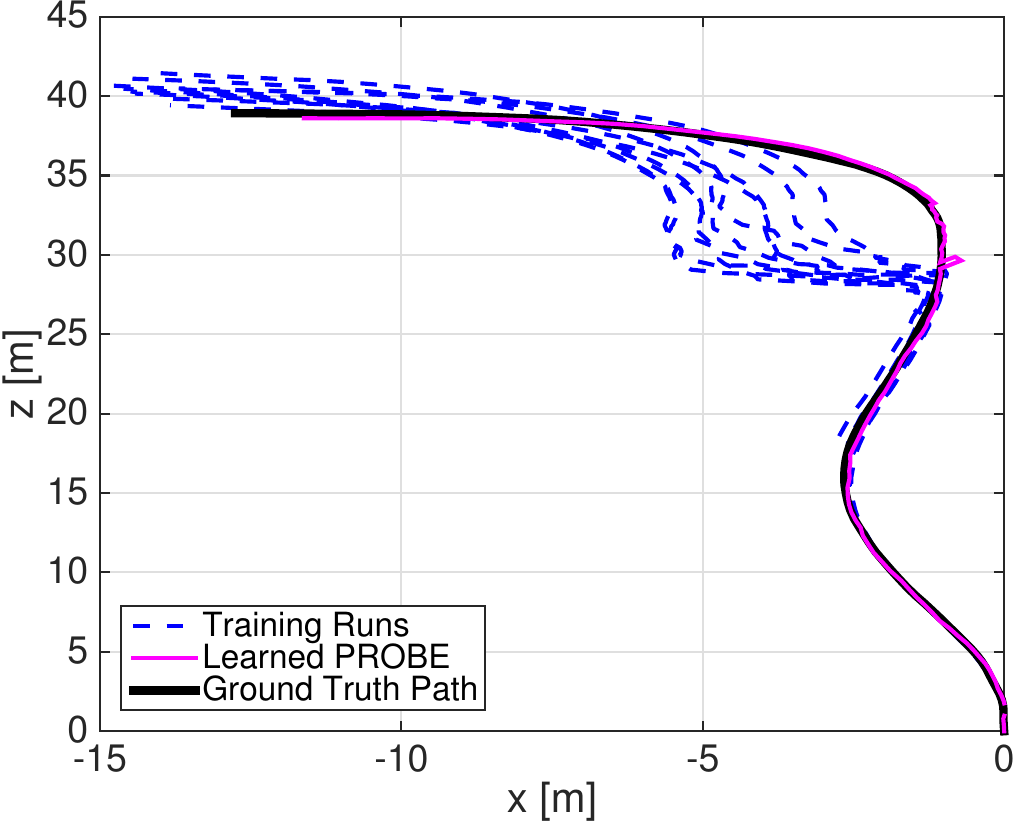}
        \put(20,30){\includegraphics[width=0.2\textwidth]{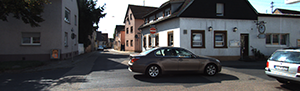}}
        \put(34, 26.5){\scriptsize Moving Vehicle}
        \thicklines
        \put(70,46){\color{gray}\vector(3,1){18}}
    \end{overpic}
        \caption{Training iterations for the Residential category, sequence \texttt{09\_26\_drive\_0046}. The left turn is particularly problematic due to a moving car that comes into view. Although none of the training runs completely remove all features from the car, the path differences are enough for the learned PROBE model to adequately reduce the influence of the car on the final motion estimate.}
    \label{fig:residential:training}
\end{figure}

Sample results of the training procedure described in Section \ref{sec:probe} are illustrated in Figure \ref{fig:residential:training} for data from the Residential category.
As ground truth is available for each image, we compute the RMSE at every frame, and only iterate over the path 10 times. Note the large variance of training run error in the sharp turn in Figure \ref{fig:residential:training}, caused by a car that drives through the camera's field of view. PROBE is able to distinguish features on the car and adequately reduce their influence with the final learned model.

\begin{figure}
    \centering
    \begin{overpic}[width=0.4\textwidth]{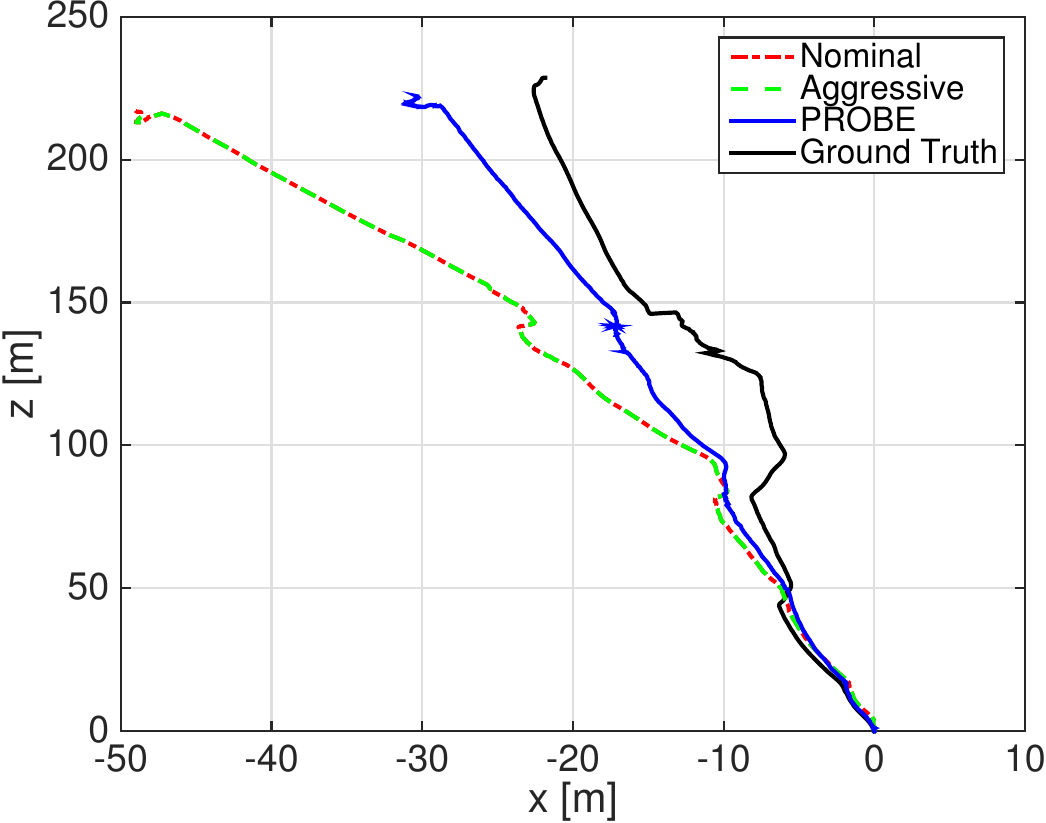}
        \put(20,21){\includegraphics[width=0.15\textwidth]{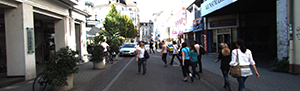}}
        \put(31,17.5){\scriptsize Pedestrians}
         \thicklines
         \put(58,30){\color{gray}\vector(1,{0.2}){10}}
    \end{overpic}
    \caption{A 234 m KITTI test run in the City category, sequence \texttt{09\_29\_drive\_0071} containing numerous pedestrians and dramatic lighting changes. PROBE is able to produce more accurate navigation estimates than even an aggressive RANSAC routine.}
    \label{fig:city:test}
\end{figure}

\begin{figure}
    \centering
    \begin{overpic}[width=0.4\textwidth]{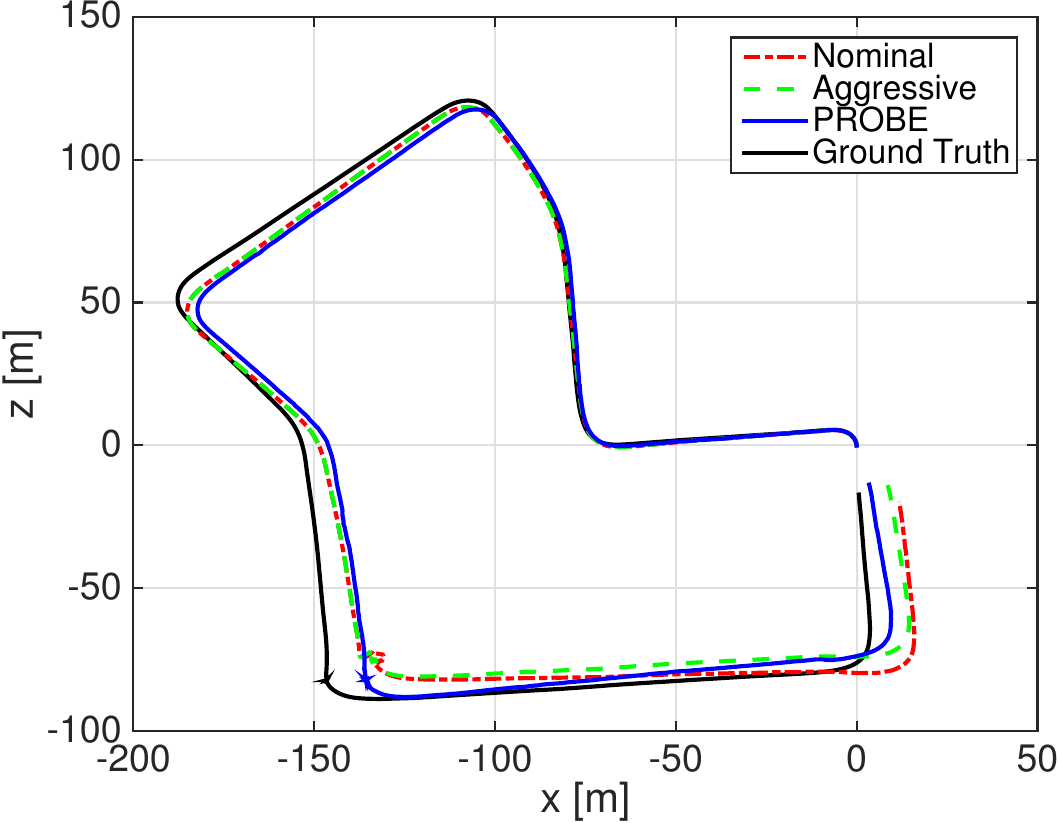}
        \put(39,21){\includegraphics[width=0.15\textwidth]{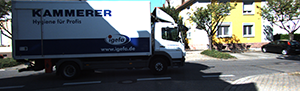}}
        \put(43,17.5){\scriptsize Vehicle with Shadow}
         \thicklines
   		  \put(40,20){\color{gray}\vector(-2,-2){3.5}}
    \end{overpic}
    \caption{A 667 m test run in the Residential category, sequence \texttt{09\_30\_drive\_0027}. PROBE is better able to deal with a static portion when a large shadow and moving vehicle cross the field of view of the camera.}
    \label{fig:residential:test}
\end{figure}

\begin{figure}
    \centering
    \begin{overpic}[width=0.4\textwidth]{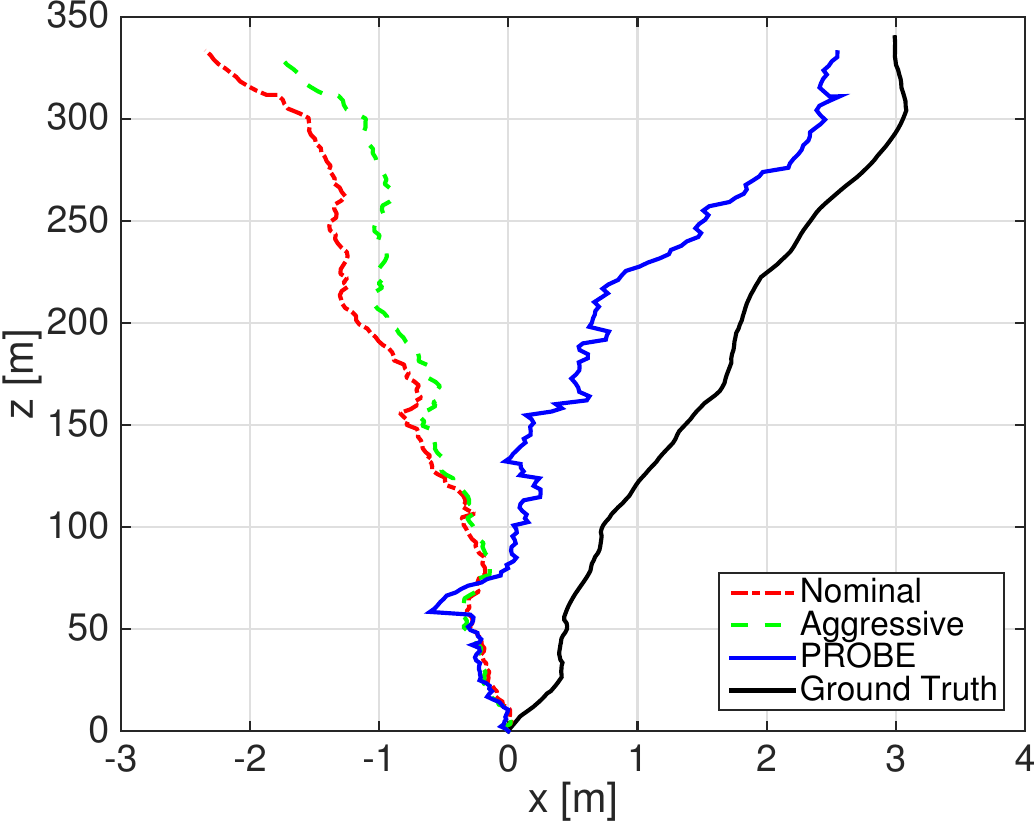}
        \put(36,65){\includegraphics[width=0.14\textwidth]{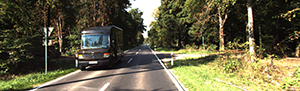}}
        \put(43,61.5){\scriptsize Moving Vehicle}
        \thicklines
        \put(48,60){\color{gray}\vector(0,-1){35}}
    \end{overpic}        
    \caption{A 440 m test run in the Road category, sequence \texttt{09\_26\_drive\_0027}. PROBE is able to better predict a large moving vehicle, and extract higher quality features from dense foliage. Note the difference in scale between the two axes.}
    \label{fig:road:test}
\end{figure}

\subsection{Evaluation}
To evaluate PROBE, we run the nominal VINS (cf. Section  \ref{sec:vins}) on a given test trial, tune the RANSAC threshold to achieve reasonable translation error ($< 5\%$ final drift), then repeat the trial with the aggressive RANSAC procedure. Finally, we run VINS again, this time disabling RANSAC completely and applying our trained PROBE model (with pre-processing) to each observed feature.  Table \ref{table:kitti_data} compares the performance of each trained PROBE model to that of the nominal and aggressive-RANSAC VINS. In the best case, PROBE achieves a final translational error norm of less than half that of both reference VINS. Figures \ref{fig:city:test}, \ref{fig:residential:test}, and \ref{fig:road:test} reveal problematic sections of the KITTI dataset where PROBE is able to significantly improve upon the performance of both reference VINS. Moving vehicles and pedestrians are the most obvious sources of error that PROBE is able to identify. The datasets also included more subtle effects such as motion blur (notable at the edge of images), slowly swaying vegetation, and shadows.

\begin{figure}
    \centering
        \begin{overpic}[width=0.4\textwidth]{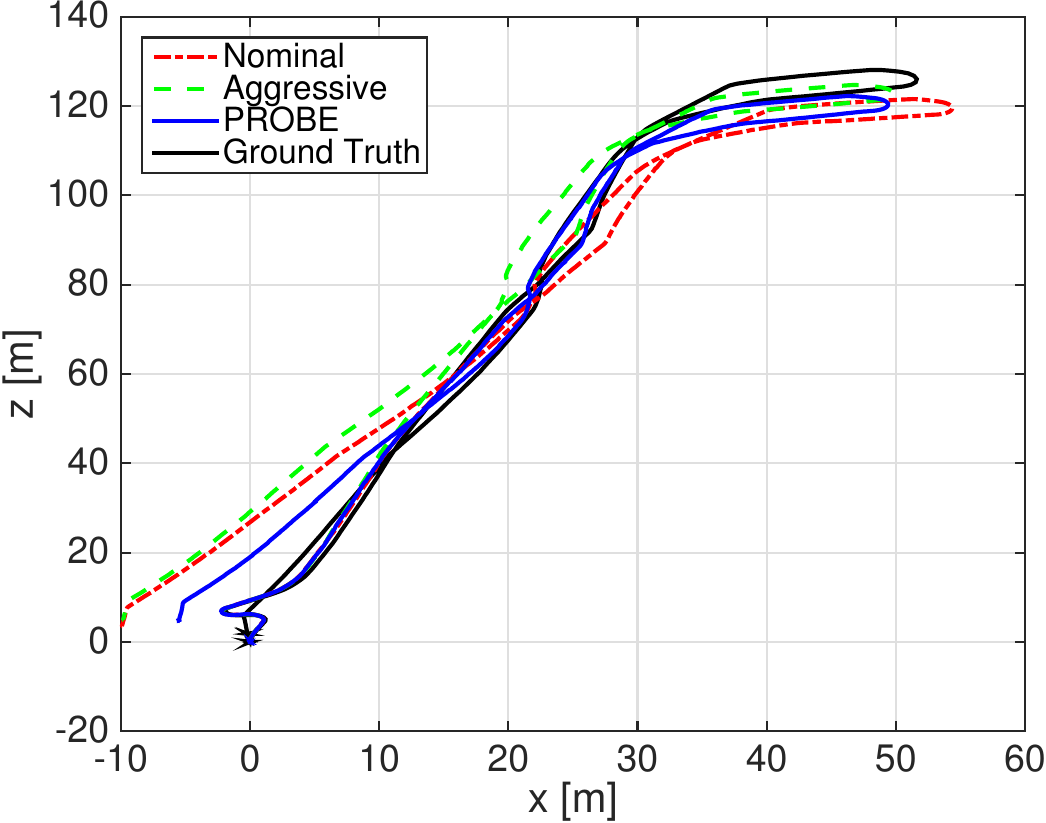}
            \put(56,10){\includegraphics[width=0.16\textwidth]{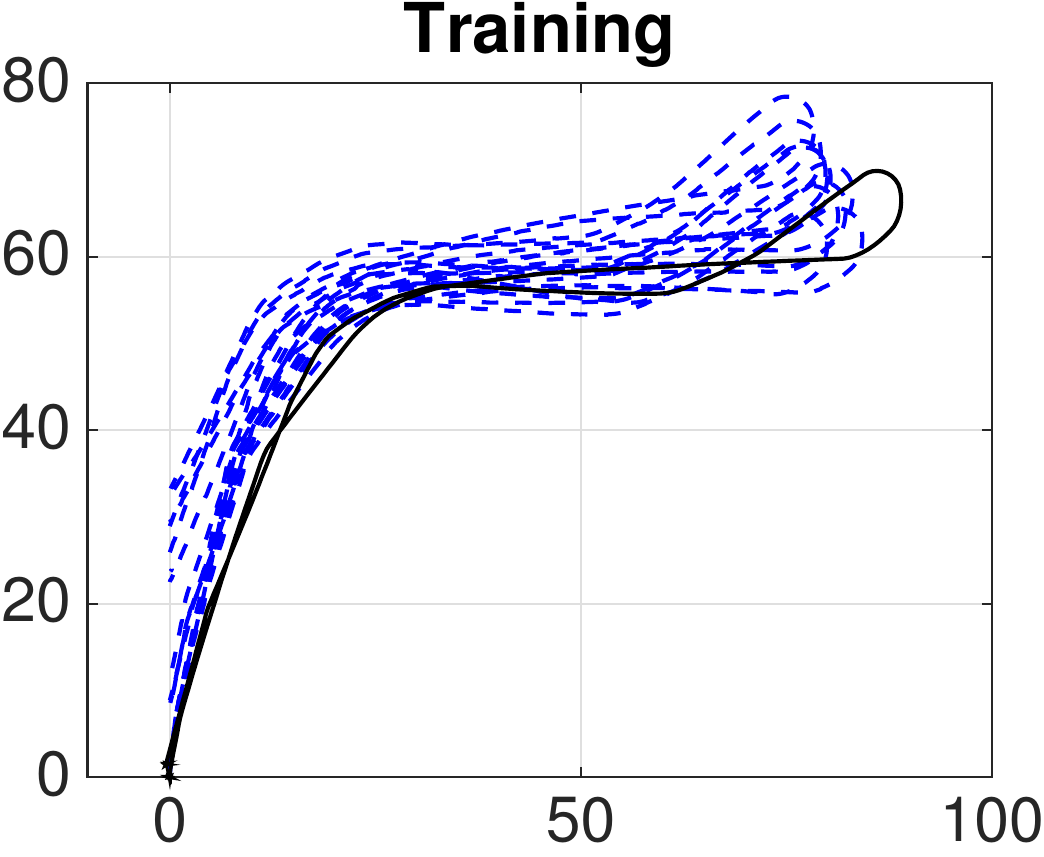}}
        \end{overpic}
    \caption{Test and training trials in the outdoor UTIAS environment. The rover traverses a snowy landscape and people walk though the field of view.}
    \label{fig:husky}
\end{figure}

\begin{figure}
    \centering
        \begin{overpic}[width=0.4\textwidth]{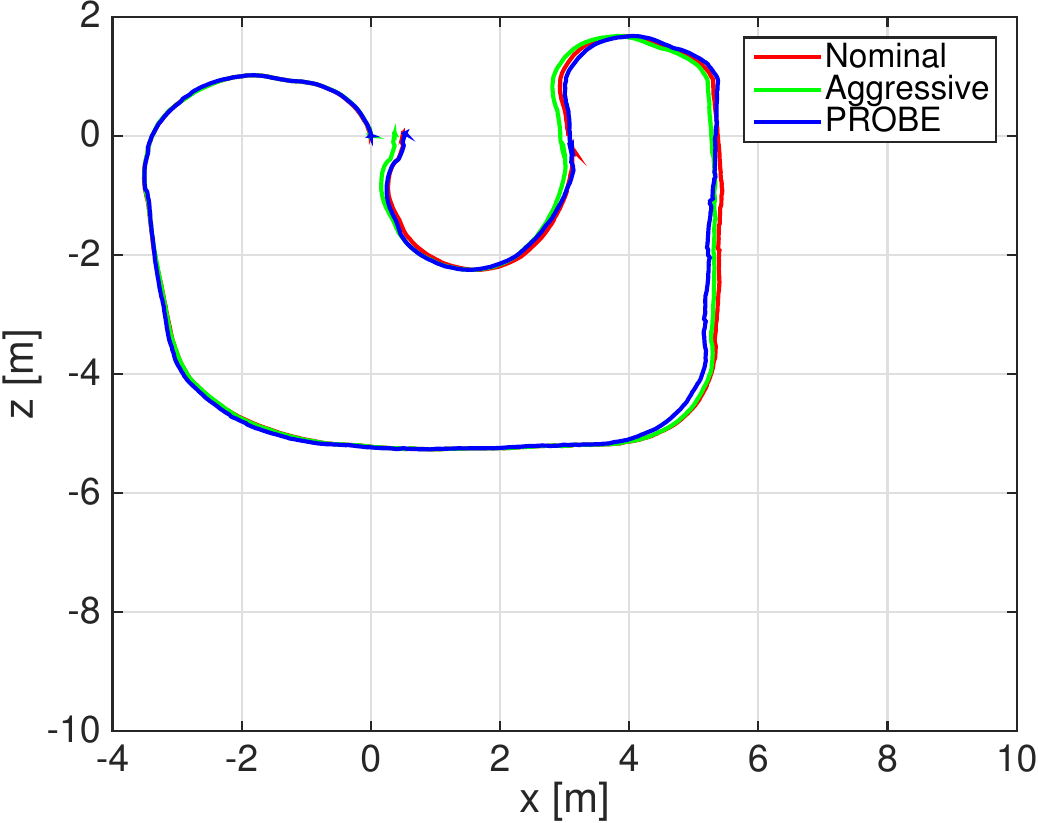}
            \put(15,12){\includegraphics[width=0.10\textwidth]{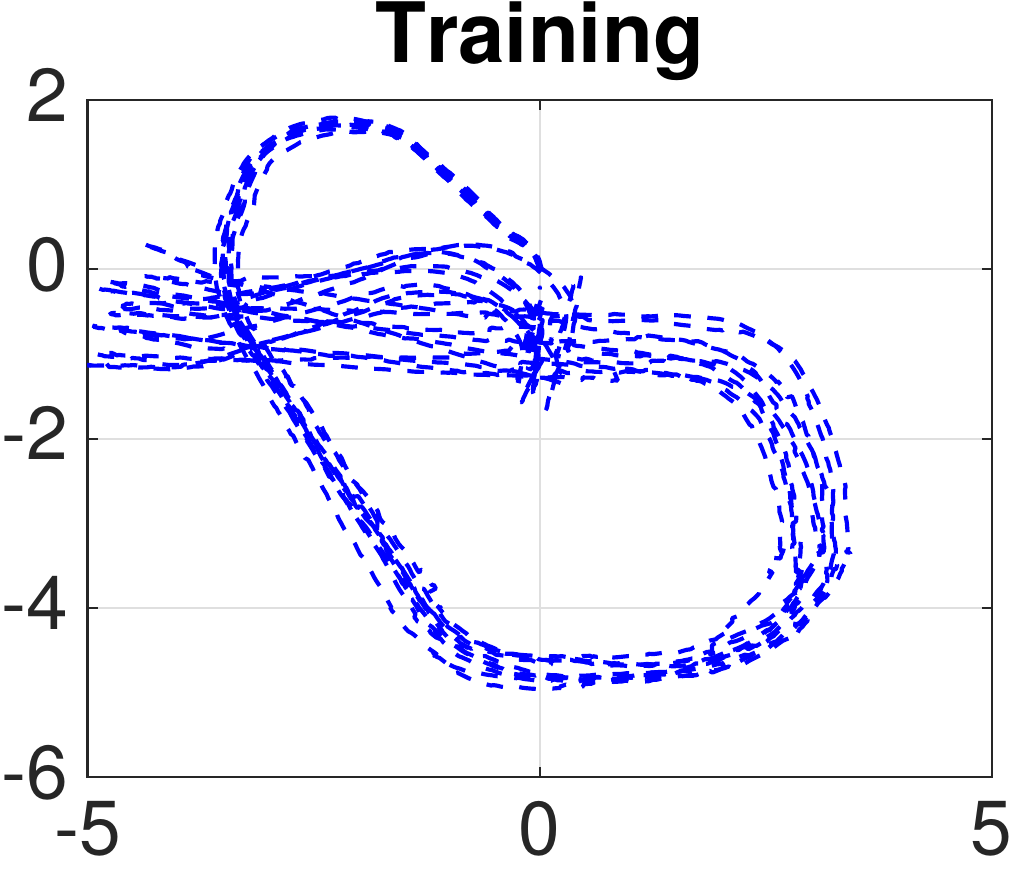}}
            \put(60,10){\includegraphics[width=0.14\textwidth]{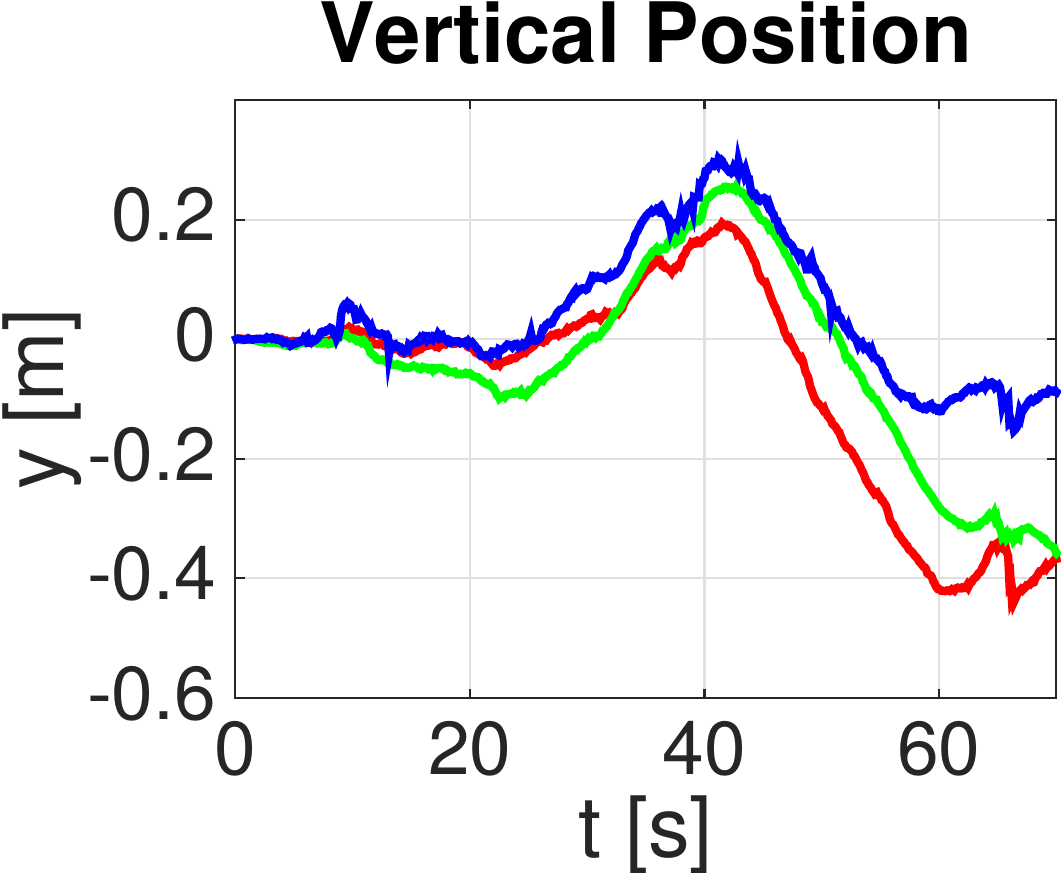}}
        \end{overpic}
    \caption{A 32.8 m test run in our indoor UTIAS dataset. The subplot in the bottom-right corner of the figure shows that PROBE reduces drift in the vertical (y) direction more than the nominal VINS and aggressive RANSAC.}
    \label{fig:pioneer}
    \vspace{-1em}
\end{figure}

%%%%%%%%%%%%%%%%%%%%%%%%%%%%%%%%%%%%%%%%%%%%%%%%%%%%%%%%%%%%%%%%%%%%%%%%%%%%%%%%
%%%%%%%%%%%%%%%%%%%%%%%%%%%%%%%%%%%%%%%%%%%%%%%%%%%%%%%%%%%%%%%%%%%%%%%%%%%%%%%%
\section{DISCUSSION} \label{sec:discussion}
In most of the datasets we evaluated, PROBE performed as well as or better than a standard RANSAC routine in reducing the influence of deleterious features. 
In particular, PROBE has proven to be more robust than aggressive RANSAC in datasets that exhibit visual effects such as shadows, large moving objects, and self-similar textures.
Often, PROBE can produce more accurate navigation estimates  by intelligently weighting measurements to reflect their quality while still exploiting the information contained in low-quality measurements. 
In this sense, PROBE can be thought of as a soft outlier classification scheme, while RANSAC rejects outliers based on binary classification.

Like with other machine learning approaches, a good training dataset is essential to producing an accurate and generalizable model. PROBE is flexible enough that it can be taught with varying frequency of ground truth data. For instance, in the outdoor dataset collected at  UTIAS, GPS measurements were available at only 1 Hz, and PROBE was trained by evaluating the ARMSE over the entire path. For the indoor dataset, no ground truth was available and the model was learned by computing the loop closure error between the start and end points of the training path.

%%%%%%%%%%%%%%%%%%%%%%%%%%%%%%%%%%%%%%%%%%%%%%%%%%%%%%%%%%%%%%%%%%%%%%%%%%%%%%%%
%%%%%%%%%%%%%%%%%%%%%%%%%%%%%%%%%%%%%%%%%%%%%%%%%%%%%%%%%%%%%%%%%%%%%%%%%%%%%%%%
\section{CONCLUSIONS} \label{sec:conclusions}
In this work, we presented PROBE, a novel method for predicting the quality of visual features within complex, dynamic environments. By using training data to learn a mapping from a predefined space of visual-inertial predictors to a scalar weight, we can adjust the influence of individual visual features on the final navigation estimate. PROBE can be used in place of traditional outlier rejection techniques such as RANSAC, or combined with them to more intelligently weight inlier measurements.

We explored a variety of potential predictors, and validated our technique using a visual-inertial navigation system on over 4 km of data from the KITTI dataset and 700 m of indoor and outdoor data collected at the University of Toronto Institute for Aerospace Studies.
Our results show that PROBE outperforms RANSAC-based binary outlier rejection in many environments, even with only sparse ground truth available during the training step.

In future work, we plan to examine a broader set of predictors, and extend the training procedure to incorporate online learning using intermittent ground truth measurements or loop closures detected by a place recognition thread. Further, we are interested in analyzing the amount of training data required for a given improvement in navigation accuracy, and in investigating PROBE's effect on estimator consistency.

%%%%%%%%%%%%%%%%%%%%%%%%%%%%%%%%%%%%%%%%%%%%%%%%%%%%%%%%%%%%%%%%%%%%%%%%%%%%%%%%
%%%%%%%%%%%%%%%%%%%%%%%%%%%%%%%%%%%%%%%%%%%%%%%%%%%%%%%%%%%%%%%%%%%%%%%%%%%%%%%%
\section*{ACKNOWLEDGEMENT}
This work was supported by the Natural Sciences and Engineering Research Council (NSERC) through the NSERC Canadian Field Robotics Network (NCFRN).
%%%%%%%%%%%%%%%%%%%%%%%%%%%%%%%%%%%%%%%%%%%%%%%%%%%%%%%%%%%%%%%%%%%%%%%%%%%%%%%%

\def\url#1{} % Get rid of url in citations -- also had to modify IEEEtran.bst
\bibliographystyle{IEEEtran}
\bibliography{probe_abbrev}

\end{document}